\pdfpageattr {/Group << /S /Transparency /I true /CS /DeviceRGB>>}

\documentclass{article} % For LaTeX2e
\usepackage{iclr2021_conference,times}

\usepackage{amsmath,amsfonts,bm}

\def\eqref#1{equation~\ref{#1}}
\def\1{\bm{1}}

\def\vr{{\bm{r}}}

\def\vy{{\bm{y}}}
\def\vz{{\bm{z}}}

\DeclareMathAlphabet{\mathsfit}{\encodingdefault}{\sfdefault}{m}{sl}
\SetMathAlphabet{\mathsfit}{bold}{\encodingdefault}{\sfdefault}{bx}{n}

\def\gF{{\mathcal{F}}}

\def\gT{{\mathcal{T}}}

\def\gY{{\mathcal{Y}}}

\newcommand{\E}{\mathbb{E}}

\usepackage{physics}
\newcommand{\C}{{\text{C}}}
\newcommand{\T}{{\text{T}}}
\newcommand{\X}{{\text{X}}}

\usepackage{hyperref}
\usepackage{url}
\usepackage{blindtext}
\usepackage{todonotes}
\usepackage{xfrac}
\usepackage{commath}

\usepackage{caption}
\usepackage{subcaption}
\usepackage{wrapfig}
\usepackage{cancel}

\usepackage{booktabs}
\usepackage{siunitx}
\usepackage{etoolbox}
\usepackage{units}
\usepackage{multirow}

\newcommand{\ubold}{\fontseries{b}\selectfont}
\robustify\ubold

\usepackage{bbold}
\usepackage{graphicx}

\title{Meta-learning using privileged information for dynamics}

\author{Ben Day, Alexander Norcliffe, Jacob Moss \&  Pietro Li\`{o}\\
Department of Computer Science\\
University of Cambridge\\
Cambridge, United Kingdom\\
\texttt{\{bjd39, alin2, jm2311, pl219\}@cam.ac.uk}
}

\iclrfinalcopy % Uncomment for camera-ready version, but NOT for submission.
\begin{document}

\maketitle

\begin{abstract}
Neural ODE Processes approach the problem of meta-learning for dynamics using a latent variable model, which permits a flexible aggregation of contextual information.
This flexibility is inherited from the Neural Process framework and allows the model to aggregate sets of context observations of arbitrary size into a fixed-length representation.
In the physical sciences, we often have access to structured knowledge in addition to raw observations of a system, such as the value of a conserved quantity or a description of an understood component.
Taking advantage of the aggregation flexibility, we extend the Neural ODE Process model to use additional information within the Learning Using Privileged Information setting, and we validate our extension with experiments showing improved accuracy and calibration on simulated dynamics tasks.
\end{abstract}

\section{Introduction \& background}
Learning using privileged information (LUPI) is a machine learning paradigm where we have access to additional information during training that may not be available at test time \citep{vapnik2009new,vapnik2015learning}. Typically this information is higher quality in some way, often it conveys some expert understanding we have about the system being modelled. To paraphrase an example of \cite{vapnik2009new}, we might have access to extensive patient records associated with biopsy scans in a training set but wish to deploy our trained model `on-the-front-line' where records are incomplete or unavailable. We would like to improve the model by using this information without coming to rely on it; to provide an accurate prognosis without human help. An analogy can be made to the role of a teacher. Besides providing corrections, skilled teachers accelerate the understanding of their students through explanations and insights. This is important in the context of learning about physics, where a new perspective often provides traction with difficult problems.

In this work we propose a method for conveying such insight in the case of modelling dynamics. \textbf{Our main contribution} is a new training-mode architecture that allows privileged information to guide learning, that results in more accurate predictions and better calibrated uncertainty estimation.

\textbf{Neural Processes.} Deep neural networks are excellent function approximators that are cheap to evaluate and straightforward to train, but typically only provide point estimates and require retraining to make use of information gained at test time. However, meta-learning (occasionally, \textit{`learning-to-learn'}) with neural networks is enjoying a resurgence in popularity as an answer to the serialised learning question, and has been applied in a range of domains \citep{hospedales2020meta}. Gaussian Processes have markedly different advantages, handling uncertainty in a principled way and adapting to new data at test time \citep{rasmussen2003gaussian}, at the cost of computational expense. Neural Processes (NP) aim to combine the best of both by learning to model a distribution over functions, framing the meta-learning problem as a latent variable model \citep{garnelo2018conditional,garnelo2018neural}.

\textbf{Neural ODEs.} Dynamical systems are a fundamental object of study in physics and are often most elegantly described using ordinary differential equations (ODE).
% Natural phenomena ranging from populations in an ecological network to bouncing balls are well described in this way.
Neural ODEs (NODE) \citep{chen2018neural} combine the representation learning capabilities of neural networks with an ODE structure to allow ODEs to be learned directly from observational data. Follow-up works have made improvements to expressivity \citep{augmented2019}, investigated time series modelling \citep{kidger2020neural,norcliffe2020second}, and tackled adaptability and uncertainty estimation using Bayesian neural networks \citep{ode2vae2019}.

\textbf{Neural ODE Processes.} An alternative to the work of \cite{ode2vae2019} for modelling uncertainty in dynamics is the Neural ODE Process (NDP) which uses a neural process derived formulation to learn a distribution over dynamics \citep{norcliffe2021neural}. These models are able to fast-adapt to new data points at test time, unlike vanilla-NODEs, whilst inheriting superior time-series modelling as compared to vanilla-NPs. The key components of the NDP are an observation encoder, representation aggregator, latent ODE, and decoder.

\begin{figure}
    \centering
    \includegraphics[trim = 40 120 900 40, clip=true, width=\textwidth]{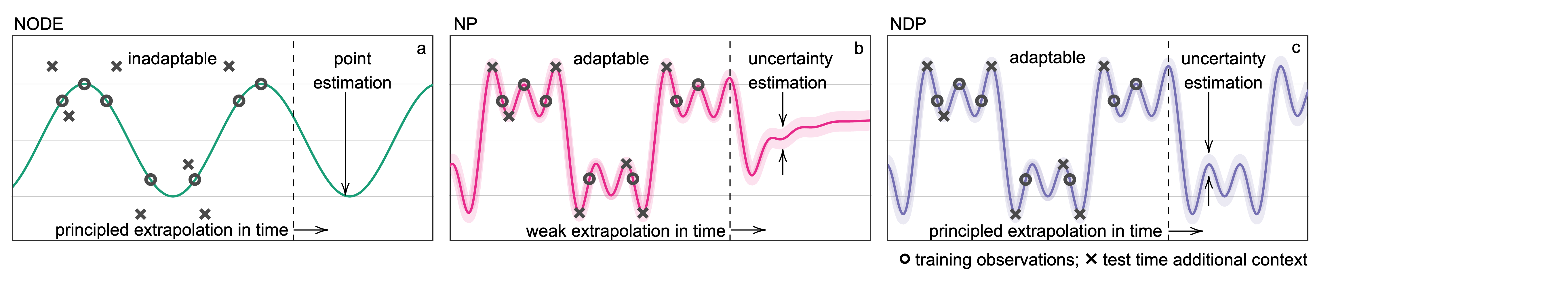}
    \caption{Comparing relevant abilities of NODEs, NPs, and NDPs. (a) NODEs explicitly learn dynamics, and can extrapolate well in time as a result, but cannot adapt to new information without retraining. (b) NPs offer uncertainty estimation and can adapt to new information but do not have a principled way to extrapolate in time. (c) NDPs combine these abilities in a single model.}
    \label{fig:comparison}
  \vspace{-12pt}
\end{figure}

\textbf{Learning using privileged information.} The LUPI framework, introduced by \cite{vapnik2009new} and expanded by \cite{vapnik2015learning}, formalises a learning setup in which a teacher is able to provide the student learner with structured explanations, comments, comparisons, etc. beyond direct supervision. \citet{hernandez2014mind} show that privileged information can be used effectively by GP classifiers, whilst \citet{lambert2018deep} apply LUPI to deep neural networks by setting the dropout rate to be a function of the privileged information.

\section{Learning using privileged information with NDPs}
\label{sec:lupi-ndp}
\textbf{Problem statement.} Formally, our setting closely matches that of the NDP: we consider modelling random functions over time, $F: \gT \to \gY$, where $F$ has distribution $D$ induced by a second distribution, $D'$, over some underlying dynamics\footnote{If the dynamics manifest directly in observation space (i.e. they are not latent), $D$ and $D'$ coincide.}. We are provided with a set of labelled samples from an instantiation $\gF$ of $F$ referred to as the context set, indexed by $I_\C$ and denoted $\C = \{(t_i^\C, \vy_i^\C) \}_{i \in I_\C}$. The task is to predict the values $\{\vy_j^\T\}_{j \in I_\T}$ taken by $\gF$ at a set of target times $\{t_j^\T\}_{j \in I_\T}$, indexed by $I_\T$, which together form the target set, $\T = \{(t_j^\T,\vy_j^\T)\}_{j \in I_\T}$.

\begin{figure}
    \centering
    \includegraphics[trim = 0 10 324 10, clip=true, width=\textwidth]{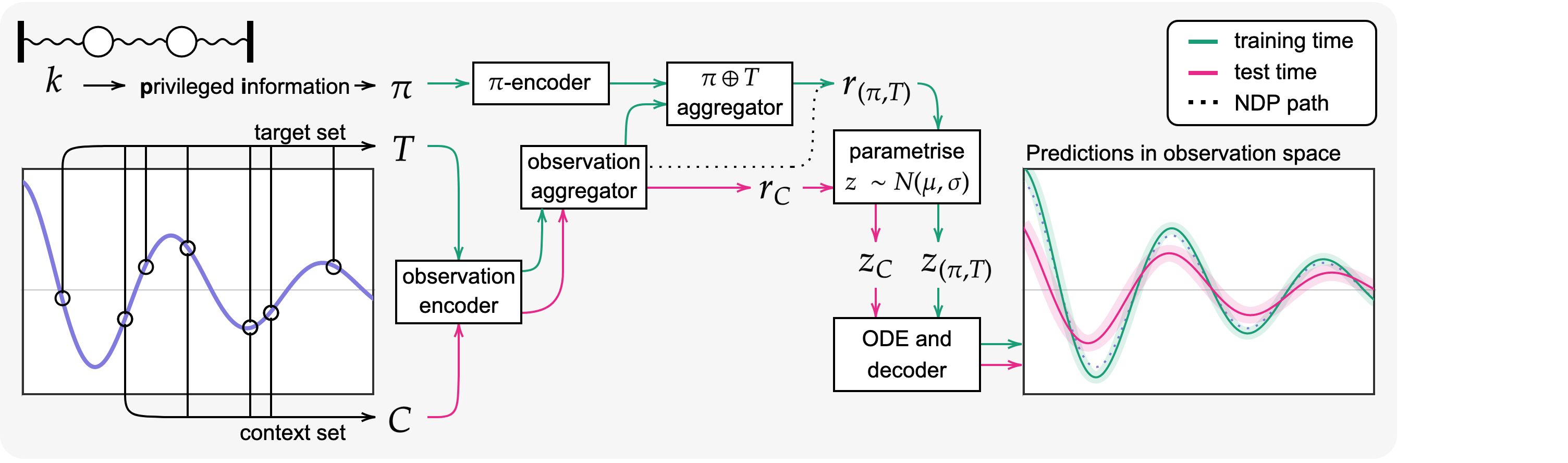}
    \caption{LUPI-NDP computational diagram. The LUPI training-mode is shown in green, with a dotted line for the vanilla-NDP path (no privileged information). Pink shows the evaluation-mode path, that matches the NDP evaluation-mode exactly. Observations are first encoded then aggregated. Privileged information, in this case the spring stiffness, is separately encoded and, during training, combined with the aggregated observations representation. As in NPs and NDPs, the representations parametrise the global latent variable $\vz$, which is then used to condition the decoder. Typically, as in this cartoon, the privileged information enables highly accurate predictions.}
    \label{fig:model_overview}
      \vspace{-12pt}
\end{figure}

To ensure the model is able to learn the underlying distribution over dynamics, and how this manifests as a distribution over functions, training assumes access to a set of time-series sampled from $F$. Diverging from the NDP setting, during training the model also has access to privileged information relating to each instantiation $\gF$ of $F$, $\boldsymbol{\pi}_\gF$. The privileged information could be some physical property or conserved quantity of the system, such as the spring stiffness, as in Figure \ref{fig:model_overview}. At test time no privileged information is provided and there is no difference from the vanilla-NDP setting.

\textbf{Model overview.} The differences introduced by the LUPI framework affect the training procedure, introducing a second source of information with which to form the global latent variable $\vz$. To incorporate the privileged information within the LUPI-NDP we introduce a second encoder to produce a representation $\vr_{\pi} = f_e^{\pi}(\boldsymbol{\pi})$, where $f_e^{\pi}$ is parametrised as a fully-connected neural network. During training, an additional aggregation step is introduced to combine the observations derived representation, $\vr_{o}$, with that of the privileged information to form a global representation, $\vr_{\textup{train}} = g(\vr_{o},\vr_{\pi})$. Practically, as the test-time global representation is that formed using only the observations, i.e. $\vr_{\textup{test}} = \vr_o$, we choose to parametrise $g$ as a residual network \citep{he2016deep}, that is, $g(\vr_{o},\vr_{\pi}) = \vr_{o} + g'(\vr_{o},\vr_{\pi})$, such that the privileged information is explicitly used as a correction term. At test time, the model we propose is equivalent to a vanilla-NDP.

\textbf{Learning and Inference.} The true posterior is intractable and, as is the NP custom, the model is trained using amortised variational inference. We select an objective that reflects the intended test time behaviour of making predictions based solely on the context set, given by
\begin{equation}
\label{eq:loss_fn}
    \log p(\vy_{j \in I_\T} | t_{j \in I_\T}, \C) \geq  \E_{q(z|\T, \boldsymbol{\pi})}\Bigg[\log \frac{q'(\vz|\C)}{q(\vz|\T,\boldsymbol{\pi})} + \sum_{i \in I_\T} \log p(\vy_i | \vz, t_i) \Bigg],
\end{equation}
with variational posterior $q$ (a derivation is provided in Appendix \ref{app:derivation}). Note that during training $\C$ is a subset of $\T$. The original NDP model is recovered by setting $\boldsymbol{\pi}$ to $\textup{null}$ (dropping it).

\section{Experiments}
We compare our proposed LUPI-NDP model with a vanilla-NDP that does not make use of the privileged information and, as there are no architectural changes, at test time the models are distinguishable only by the value of their weights. As explained in Section \ref{sec:lupi-ndp}, both models are NDPs, so we refer to them as LUPI and NoPI (no-privileged-information) for clarity. Full model and training details are provided in Appendix \ref{app:details}, code at  \href{https://github.com/bjd39/lupi-ndp}{\texttt{github.com/bjd39/lupi-ndp}}.

\begin{wrapfigure}{r}{0.30\textwidth}
    \vspace{-25pt}
    \begin{center}
        \includegraphics[width=0.28\textwidth]{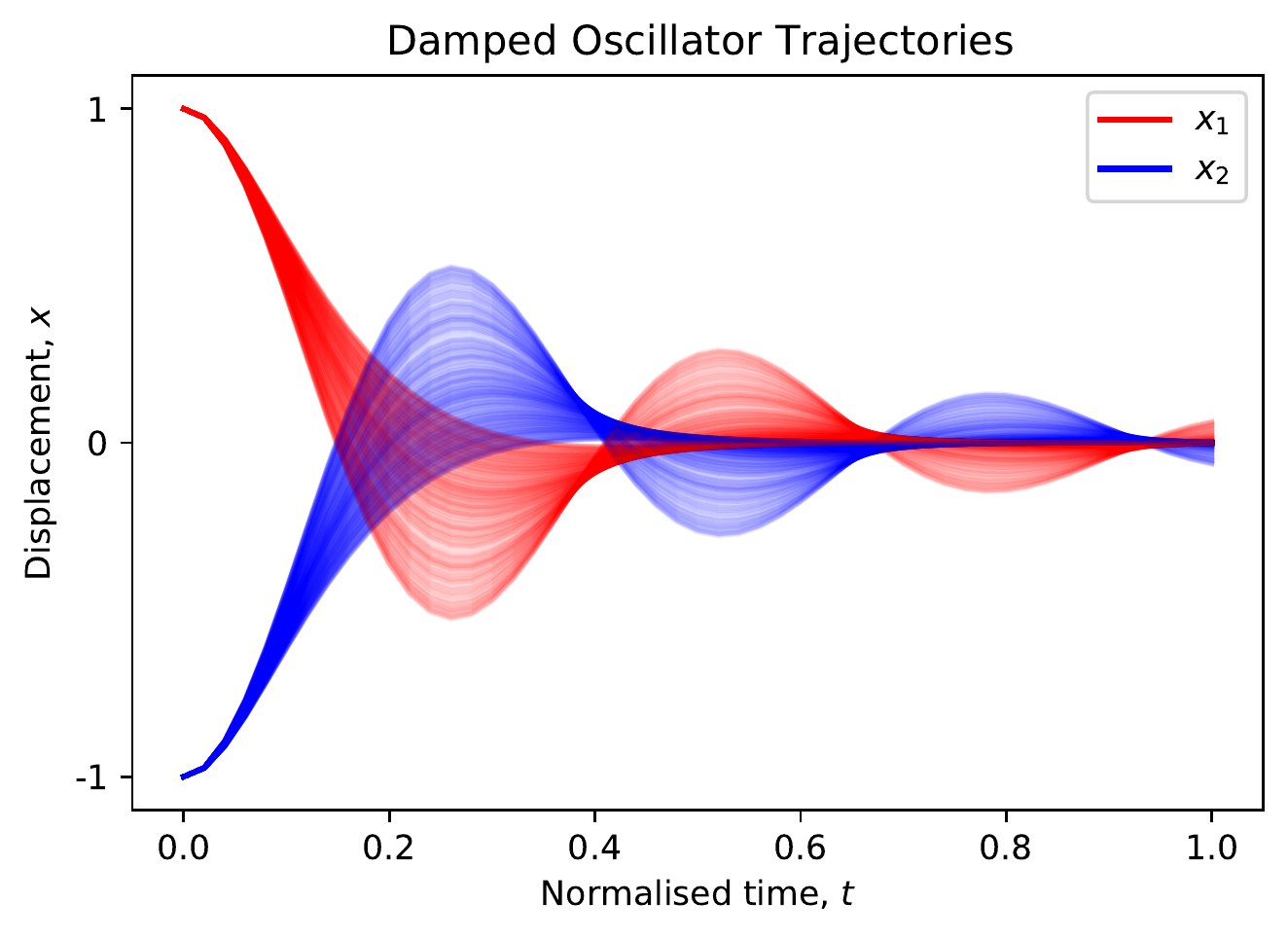}
        \includegraphics[width=0.28\textwidth]{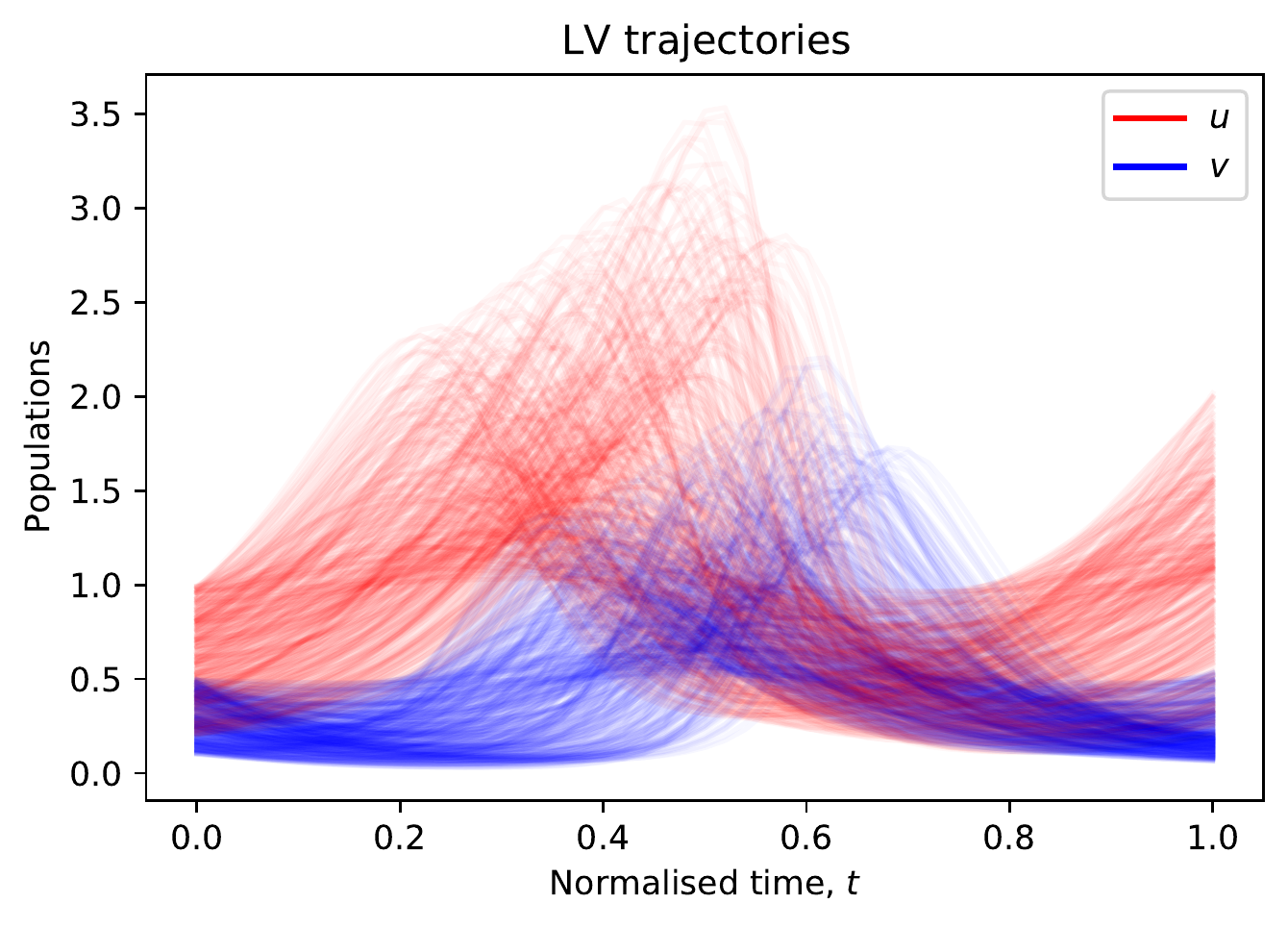}
    \end{center}
  \vspace{-15pt}
  \caption{Induced dynamics.}
  \label{fig:induced}
\end{wrapfigure}

\textbf{Metrics.} Besides providing high quality predictions, NDPs, and NPs generally, are interesting because the way they approach the meta-learning problem (as a latent variable model) produces uncertainty estimates. We measure the quality of these estimates by the calibration error and sharpness. \textit{Calibration} measures the degree to which uncertainty estimates are commensurate with residuals: if the model estimates an outcome to happen one time in every ten, does it actually occur that frequently? \textit{Sharpness} is simply how low the uncertainty estimates are: between equally well-calibrated models we should favour the sharper model as being more informative. Practically speaking, we value accuracy over calibration over sharpness. We follow the definitions of \cite{kuleshov2018accurate} for the calibration error and sharpness, detailed in Appendix \ref{app:calibration}.

\textbf{Damped coupled oscillators.} We first consider modelling a system of two masses attached by identical springs in series between parallel walls, described by the second-order differential equations
\begin{equation}
    m_1 \ddot{x_1} = (x_2 - 2x_1)k - c\dot{x_1} \quad , \quad m_2 \ddot{x_2} = (x_1 - 2x_2)k - c\dot{x_2},
\end{equation}
with spring and damping constants, $k$ and $c$, respectively. A distribution over dynamics is induced by sampling some parameters of the system, and this parameter forms the privileged information. Examples of the trajectories induced by sampling over the drag coefficient as $c \sim \textup{U}(0.5,2)$ are presented in Figure \ref{fig:induced}.

\textbf{Lotka-Volterra.} We investigate modelling a second two dimensional system, the Lotka-Volterra equations (L-V). The populations of `predator' and `prey' species, $v$ and $u$, are governed by
\begin{equation}
    \dot{u} = \alpha u - \beta uv \quad ; \quad \dot{v} = \delta uv - \gamma v.
\end{equation}
In these examples we use the values $\alpha = \nicefrac{2}{3}, \beta = \nicefrac{4}{3}, \gamma = 1, \delta = 1$. To produce a range of dynamics we uniformly sample different initial populations $u_{0} \sim \textup{U}(0.2,1), v_{0} \sim \textup{U}(0.1,0.5)$ as shown in Figure \ref{fig:induced}. For these equations $V = \delta u -\gamma \ln(u) + \beta v -\alpha \ln(v)$ is a conserved quantity, and is provided as privileged information.

\begin{figure}
     \centering
     \begin{subfigure}[b]{0.32\textwidth}
         \centering
         \includegraphics[trim = 8 5 8 21, clip=true, width=\textwidth]{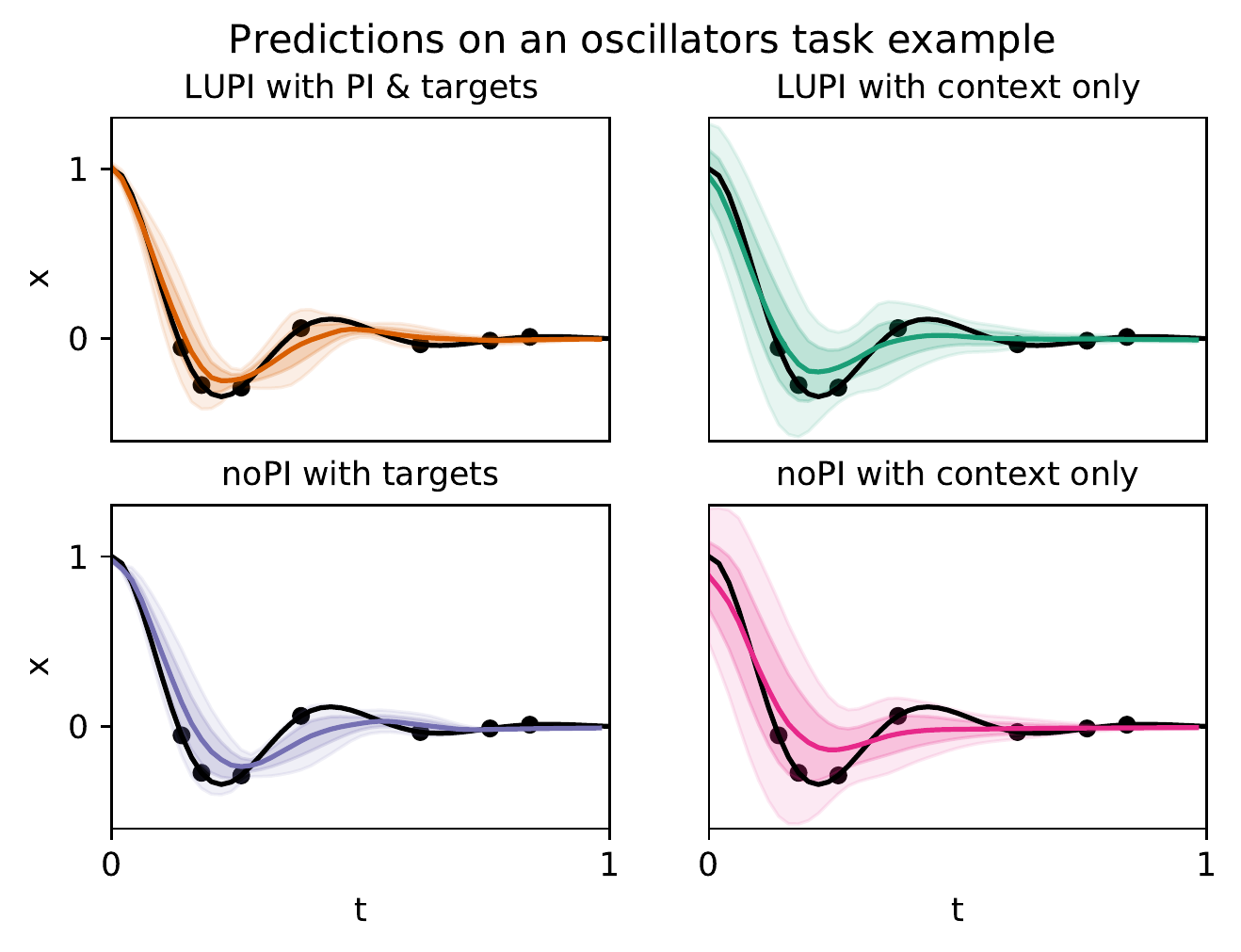}
         \caption{Example predictions}
         \label{fig:predictions}
     \end{subfigure}
     \hfill
     \begin{subfigure}[b]{0.32\textwidth}
         \centering
         \includegraphics[trim =  5 5 5 21, clip=true, width=\textwidth]{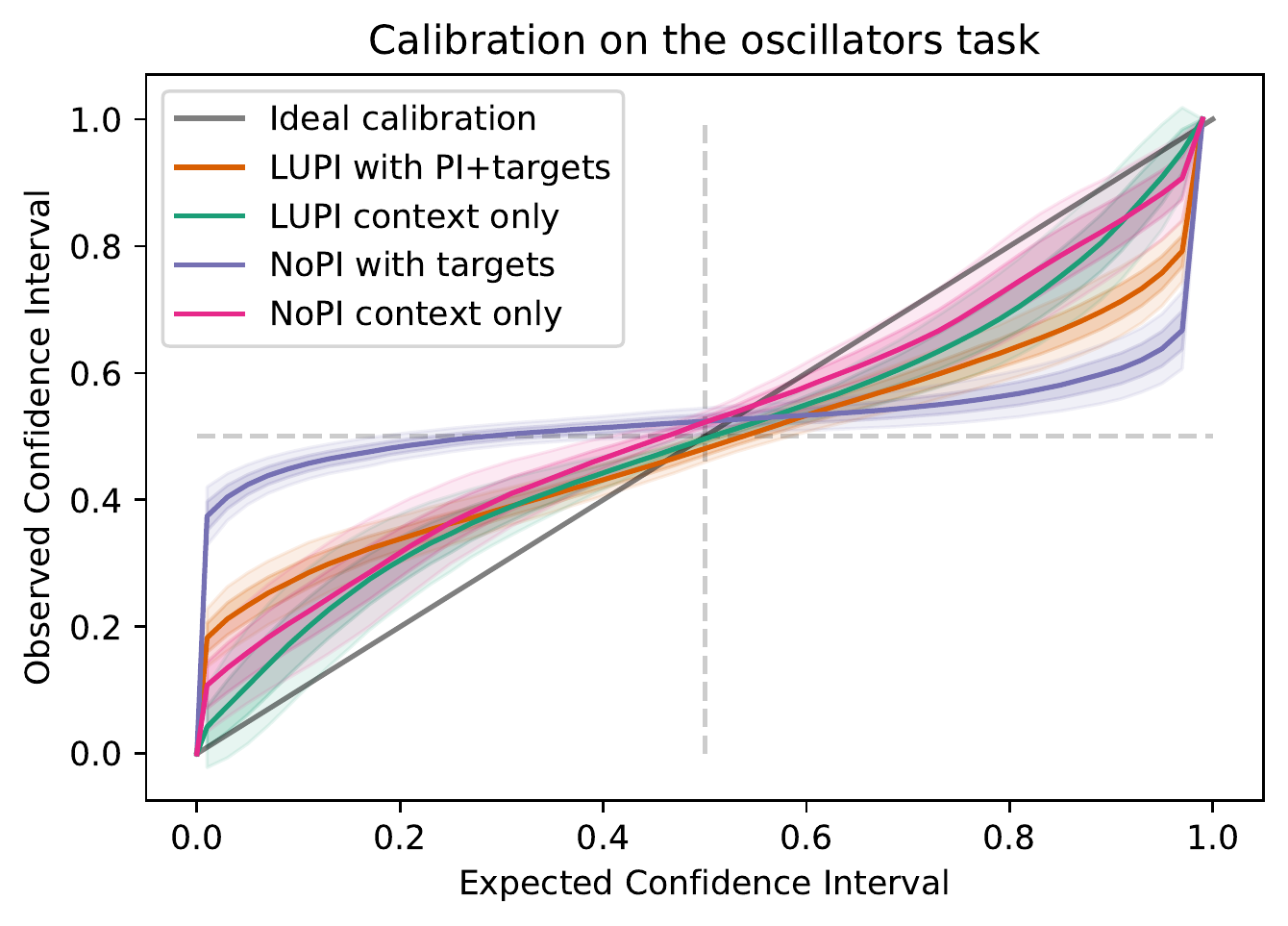}
         \caption{Calibration}
         \label{fig:calibration}
     \end{subfigure}
     \begin{subfigure}[b]{0.32\textwidth}
         \centering
         \includegraphics[trim = 5 5 5 21, clip=true, width=\textwidth]{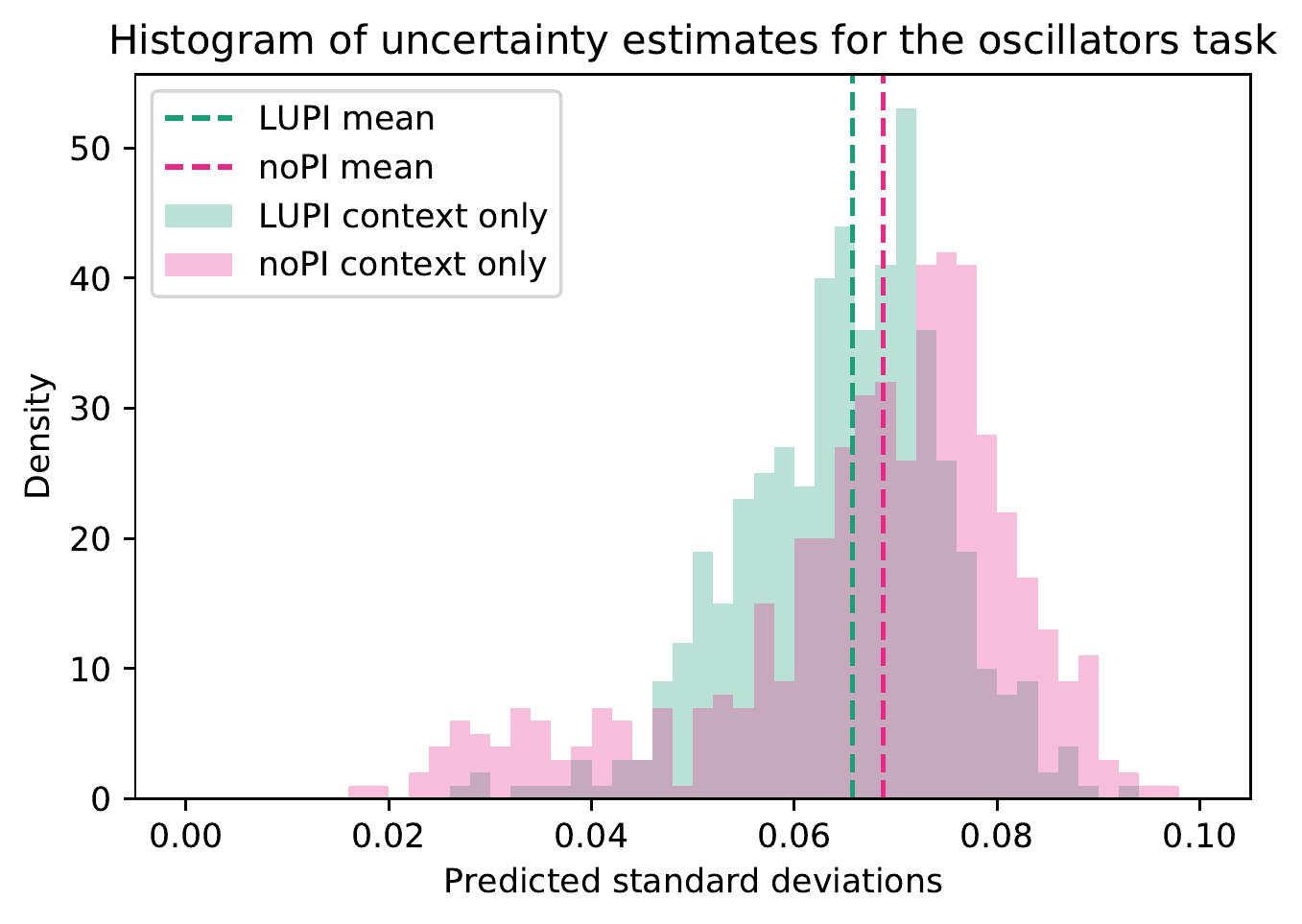}
         \caption{Sharpness}
         \label{fig:sharpness}
     \end{subfigure}
        \caption{Various comparisons of trained LUPI and NoPI models on the damped coupled oscillators task with varying stiffness, $k$. The LUPI model is more accurate, better calibrated and sharper. All the models are overconfident to some degree, but the training-mode NoPI is by far the least calibrated. Though the LUPI model is sharper, the NoPI model has greater dispersion (spread in uncertainty estimates) which would be preferable if the model were better calibrated.}
        \label{fig:three_plot_osc}
\end{figure}

\begin{table}[t]
\caption{Mean squared error (MSE) and measures of uncertainty quality for the varying-stiffness, varying-damping, and L-V tasks. Models labelled * were evaluated in the training setting, i.e. using the full target set for context and, in the case of the LUPI models, the privileged information, and are included for reference. Lower is better; the better performance in each bracket is indicated in \textbf{bold}; units are arbitrary and results should not be compared between the tasks.}
\vspace{-5pt}
\small
\label{tab:full_results}
\begin{subtable}{\textwidth}
    \centering
    \begin{tabular}{lccc|ccc}
    \toprule
    & \multicolumn{3}{c|}{Varying stiffness, $k \sim \textup{U}(0.2,1)$} & \multicolumn{3}{c}{Varying damping, $c \sim \textup{U}(0.5,2)$}
    \\
    Model  &\multicolumn{1}{c}{MSE $\downarrow$}  &\multicolumn{1}{c}{Calib. error $\downarrow$}  &\multicolumn{1}{c|}{Sharp. $\downarrow$} &\multicolumn{1}{c}{MSE $\downarrow$}  &\multicolumn{1}{c}{Calib. error $\downarrow$}  &\multicolumn{1}{c}{Sharp. $\downarrow$} 
    \\ \midrule 
    NoPI &  1.05 $\pm$ 0.05 & 0.51 $\pm$ 0.02 & 6.88 & 2.82 $\pm$ 0.29 & 0.84 $\pm$ 0.04 & 2.15\\
    LUPI  & \bf 0.93 $\pm$ \bf 0.04 & \bf 0.47 $\pm$ 0.02 & 6.57 & \bf 2.39 $\pm$ 0.09 & \bf 0.37 $\pm$ 0.02 & 4.71\\ \midrule
    NoPI* & 0.16 $\pm$ 0.02 & 2.69 $\pm$ 0.02 & 1.00 & 0.56 $\pm$ 0.02 & 1.56 $\pm$ 0.03 & 0.93\\
    LUPI* & \bf 0.06 $\pm$ 0.01 & \bf 0.91 $\pm$ 0.02 & 1.10 & \bf 0.25 $\pm$ 0.01 & \bf 0.73 $\pm$ 0.03 & 1.18 \\
    \bottomrule
    \end{tabular}
    \vspace{2pt}
\end{subtable}
\begin{subtable}{\textwidth}
    \label{tab:lv}
    \begin{center}
    \begin{tabular}{lccc}
    \toprule
    & \multicolumn{3}{c}{L-V, $u_{0} \sim \textup{U}(0.2,1), v_{0} \sim \textup{U}(0.1,0.5)$}
    \\
    Model  &\multicolumn{1}{c}{MSE $\downarrow$}  &\multicolumn{1}{c}{Calib. error $\downarrow$}  &\multicolumn{1}{c}{Sharp. $\downarrow$}  
    \\ \midrule 
    NoPI  & 6.44 $\pm$ 0.44 & 2.19 $\pm$ 0.05 &   2.23 \\
    LUPI  & \bf 1.82 $\pm$ \bf 0.13 & \bf 0.90 $\pm$ \bf 0.04 &  3.44 \\
    \midrule
    NoPI* & 5.24 $\pm$ 0.30 & 2.89 $\pm$ 0.04 &  1.37 \\
    LUPI* & \bf 0.73 $\pm$ \bf 0.02 & \bf 1.23 $\pm$ \bf 0.04 & 1.48  \\
    \bottomrule
    \end{tabular}
    \end{center}
    \vspace{-10pt}
\end{subtable}
\vspace{-5pt}
\end{table}

\section{Discussion}
Training in the LUPI setting produces significantly more accurate and better calibrated models in each task. Table \ref{tab:full_results} presents numerical results, Figure \ref{fig:three_plot_osc} provides more detail for the varying stiffness task, and further plots are provided in the Appendix. The models are mostly less sharp ($\sfrac{5}{6}$) but sharpness is a secondary measure to calibration, especially when the sharper model is less accurate, as is the case here. We would also highlight the subtle difference in the kinds of information being incorporated by the models---stiffness and damping are independent variables, whereas $V$ is a conserved quantity that arises from the dynamical system directly---and suggest this is a promising result for the wider applicability of the model. Future work could explore the effects LUPI has on generalisation, recovering estimates of the privileged information at test-time, and, as the model can be applied as-is to any Neural Process, tasks other than dynamics.

\newpage

\nocite{harrower2003colorbrewer}

\bibliography{iclr2021_conference}
\bibliographystyle{iclr2021_conference}

\newpage
\appendix

\subsection*{Acknowledgements}
We'd like to thank C\u{a}t\u{a}lina Cangea, Nikola Simidjievski and Cristian Bodnar for their valuable feedback on this work. JM is supported by a GlaxoSmithKline grant.

\section{Objective derivation}
\label{app:derivation}
At test time we want to (accurately) predict the targets $y_{j\in I_\T}$ at known times $t_{j \in I_T}$ given the context set $\C$, which means maximising $p(y_{j\in I_\T} | t_{j \in I_T}, \C)$ during training. Knowing that we want to end up with something similar to the objectives used by \cite{garnelo2018neural} and \cite{norcliffe2021neural}, i.e. something like
\begin{align*}
    \log p(\vy_{j\in I_\T} | t_{j\in I_\T}, \C) \geq \mathbb{E}_{q(z|\T)}\left[ \log\frac{q(\vz|\C)}{q(\vz|\T)}+\sum_{i\in I_\T}p(y_i|t_i,\vz)\right],
\end{align*}
we start with the marginal
\begin{align*}
    p(\vy_{j\in I_\T} | t_{j\in I_\T}, \C) = \int \dif \vz \ p(\vz|t_{j\in I_\T}, \C) p(y_{j\in I_\T} | t_{j\in I_\T}, \vz, \C).
\end{align*}
Noting $p(y_{j\in I_\T} | t_{j\in I_\T}, \vz, \C) = p(y_{j\in I_\T} | t_{j\in I_\T}, \vz)$ and $p(\vz|t_{j\in I_\T}, \C) = p(\vz|\C)$, we multiply by $q(\vz|\T,\boldsymbol{\pi})/q(\vz|\T,\boldsymbol{\pi})=1$ to get
\begin{align*}
    p(\vy_{j\in I_\T} | t_{j\in I_\T}, \C) &= \int \dif \vz \  q(\vz|\T,\boldsymbol{\pi})\frac{p(\vz|\C)}{q(\vz|\T,\boldsymbol{\pi})} p(y_{j\in I_\T} | t_{j\in I_\T}, \vz) \\
    &= \mathbb{E}_{q(\vz|\T,\boldsymbol{\pi})}\left[\frac{p(\vz|\C)}{q(\vz|\T,\boldsymbol{\pi})}p(y_{j\in I_\T} | t_{j\in I_\T}, \vz)\right].
\end{align*}
As usual, $p(\vz|\C)$ is intractable, and we approximate it with $q'(\vz|\C)=q(\vz|\C,\boldsymbol{\pi}\leftarrow\textbf{null})$. Finally, applying Jensen's inequality produces our objective
\begin{equation}
    \log p(\vy_{j \in I_\T} | t_{j \in I_\T}, \C) \geq  \E_{q(z|\T, \boldsymbol{\pi})}\Bigg[\log \frac{q'(\vz|\C)}{q(\vz|\T,\boldsymbol{\pi})} + \sum_{i \in I_\T} \log p(\vy_i | \vz, t_i) \Bigg].
\end{equation}
This objective is similar to the evidence lower-bound but better reflects the intended test time model behaviour.

\section{Architectural and training details}
\label{app:details}
The context/targets distinction is related to training and evaluation and is separate from the architecture of the model. As such, for this explanation we refer to the set of observations as $\X=\{(t_i^\X,\vy_i^\X)\}_{i \in I_\X}$ indexed by $I_\X$, for which we can substitute $\C$ or $\T$ as appropriate. We conceive of the model as consisting of
\begin{enumerate}
    \item an observation encoder mapping observations to representations $\vr_i=f_{\textup{obs}}(t_i,\vy_i)$
    \item an aggregator that combines observation representations into a fixed length representation $\vr_\X=\oplus_{i \in I_\X}(\vr_i)$
    \item a privileged information encoder mapping the privileged information to a representation $\vr_{\pi} = f_{\pi}(\boldsymbol{\pi})$
    \item a second aggregator that combines $\vr_\X$ and $\vr_\pi$, $\vr = g(\vr_\X, \vr_\pi)$, where $\vr$ and $\vr_\X$ have the same dimensionality
    \item a pair of functions, $\mu(\vr),\sigma(\vr)$, that parametrise the global latent variable $\vz$ as a function of either $\vr$ or $\vr_\X$, $\vz \sim \textup{N}(\mu(\vr),\sigma(\vr))$
    \item a function to initialise the latent ODE state from a sample from the global latent variable, $L(0)=f_\textup{init}(z')$
    \item the neural ODE derivative as a function of time, the instantaneous latent state and the global latent sample, $\dfrac{dL}{dt}=f_\textup{ODE}(L(t),z',t)$
    \item and a decoder that maps from the latent ODE state back to observation space and also depends on the sample from the global latent, $\hat{y}(t) = f_\textup{dec}(L(t),z')$
\end{enumerate}
as is the fashion, we chose to use neural networks for every parametric function, that is all but the first aggregation $\oplus$. Now we know what the parts are for, the architecture we used is
\begin{enumerate}
    \item $f_{\textup{obs}}$: a three-layer MLP with ReLU activations on the hidden layers (not the output) and hidden dimension 16
    \item $\oplus$: the mean concatenated with the LogSumExp (a smooth approximation of max)
    \item $f_{\pi}$: a three-layer MLP with ReLU activations on the hidden layers (not the output) and hidden dimension 16
    \item $g$: a ResNet with an input-to-output skip connection for $\vr_\X$ and a residual connection consisting of a three-layer MLP with ReLU activations on the hidden layers (not the output) and hidden dimension 16
    \item $\mu,\sigma$: two three-layer MLPs with ReLU activations on the hidden layers (not the output) and hidden dimension 16, with shared weights in the first two hidden layers
    \item $f_\textup{init}$: a three-layer MLP with ReLU activations on the hidden layers (not the output) and hidden dimension 16
    \item $f_{\textup{ODE}}$: a three-layer MLP with $\textup{softplus}(x) = \log{(1+\exp(x))}$ activations on the hidden layers (not the output) and hidden dimension 16
    \item $f_\textup{dec}$: a three-layer MLP with ReLU activations on the hidden layers (not the output) and hidden dimension 16.
\end{enumerate}
We did not undertake any extensive hyperparameter tuning though we can report that neither model is able to learn if the hidden dimension is set to be small ($<4$). We follow the best practices established by \cite{le2018empirical} for training NPs, and during training use a learned uncertainty estimate rather than resampling. This estimate is produced as an additional output from the decoder.

We use Adam with a learning rate of $0.001=10^{-3}$ and otherwise default PyTorch settings, and train for 100 epochs with a training set of 500 examples split 80/20 between train/validation. We do not use early stopping as we found that the models are stable at convergence i.e. they do not diverge when ``overtrained''. The test sets consist of a further 500 examples (of 100 time steps each) over which the reported evaluation metrics were calculated.

Data generators, and iPython notebooks and Google Colabs for running our experiments can be found at \url{https://github.com/bjd39/lupi-ndp}.

\section{Calibration \& sharpness for regression}
\label{app:calibration}
These descriptions and definitions borrow heavily from \cite{kuleshov2018accurate} and are included for completeness.

\subsection{Calibration}
\paragraph{Classification.} In the classification setting, we say a forecaster (making a large number of predictions) is calibrated if events that are assigned some probability occur about that frequently. Formally, a forecaster $\textup{H}$ is calibrated if
\begin{equation*}
    \frac{\sum_{t=1}^Ty_t\mathbb{1}\{\textup{H}(x_t)=p\}}{\sum_{t=1}^T\mathbb{1}\{\textup{H}(x_t)=p\}} \rightarrow p \ \forall \ p \in [0,1] \textup{ as } T \rightarrow \infty,
\end{equation*}
that is, in the long run ($T \rightarrow \infty$) predictions assigned probability $p$ ($\textup{H}(x_t)=p$) occur ($y_t = 1$) with frequency $p$.

\paragraph{Regression.} In the regression setting, the forecaster outputs a cumulative distribution function $F_t$ targeting $y_t$. Calibration here means that $y_t$ should fall within a 90\% confidence interval approximately 90\% of the time. We can formalise this using the quantile function, defined as returning the threshold value that random draws from $F_t$ would fall below $p$ of the time. For quantile function $Q_t(p)=\inf\{y:p \leq F_t(y)\}$, we define calibration to mean
\begin{align*}
    \frac{\sum_{t=1}^T\mathbb{1}\{y_t\leq Q_{t}(p)\}}{T} \rightarrow p \ \forall \ p \in [0,1] \textup{ as } T \rightarrow \infty.
\end{align*}
That is, in the long-run ($T \rightarrow \infty$) targets fall below the quantile function at $p$ ($y_t\leq Q_{t}(p)$) with frequency $p$, for any $p$.

To produce a calibration score, we consider how far the empirical calibration deviates from a perfectly calibrated model. This is measured by computing the empirical frequency at a set of confidence levels  $0\leq p_1 < p_2 < ... < p_m \leq 1$ as
\begin{equation}
    \hat{p}_j = \frac{|\{y_t | F_t(y_t) \leq p_j, t= 1, \dots, T\}|}{T}
\end{equation}
i.e. how often do the targets fall at a confidence level that is less than the threshold $p_j$, and computing the score
\begin{equation}
    \textup{cal}(F_1,y_1,...,F_t,y_t)=\sum_{j=1}^m(p_j - \hat{p}_j)^2
\end{equation}
(which is the mean-squared-error of the cumulative histogram of confidence over empirical frequency from the identity.)

\subsection{Sharpness}
A sharp forecast has confidence intervals that are tightly bound, and the sharpness score can be more straightforwardly defined as the mean variance of the cumulative distribution $F_t$,
\begin{equation*}
    \textup{sha}(F_1,\dots,F_T)=\frac{1}{T}\sum_{t=1}^T\textup{var}(F_t).
\end{equation*}

\section{Lotka-Volterra plots}
\begin{figure}
     \centering
     \begin{subfigure}[b]{0.32\textwidth}
         \centering
         \includegraphics[trim = 8 5 8 21, clip=true, width=\textwidth]{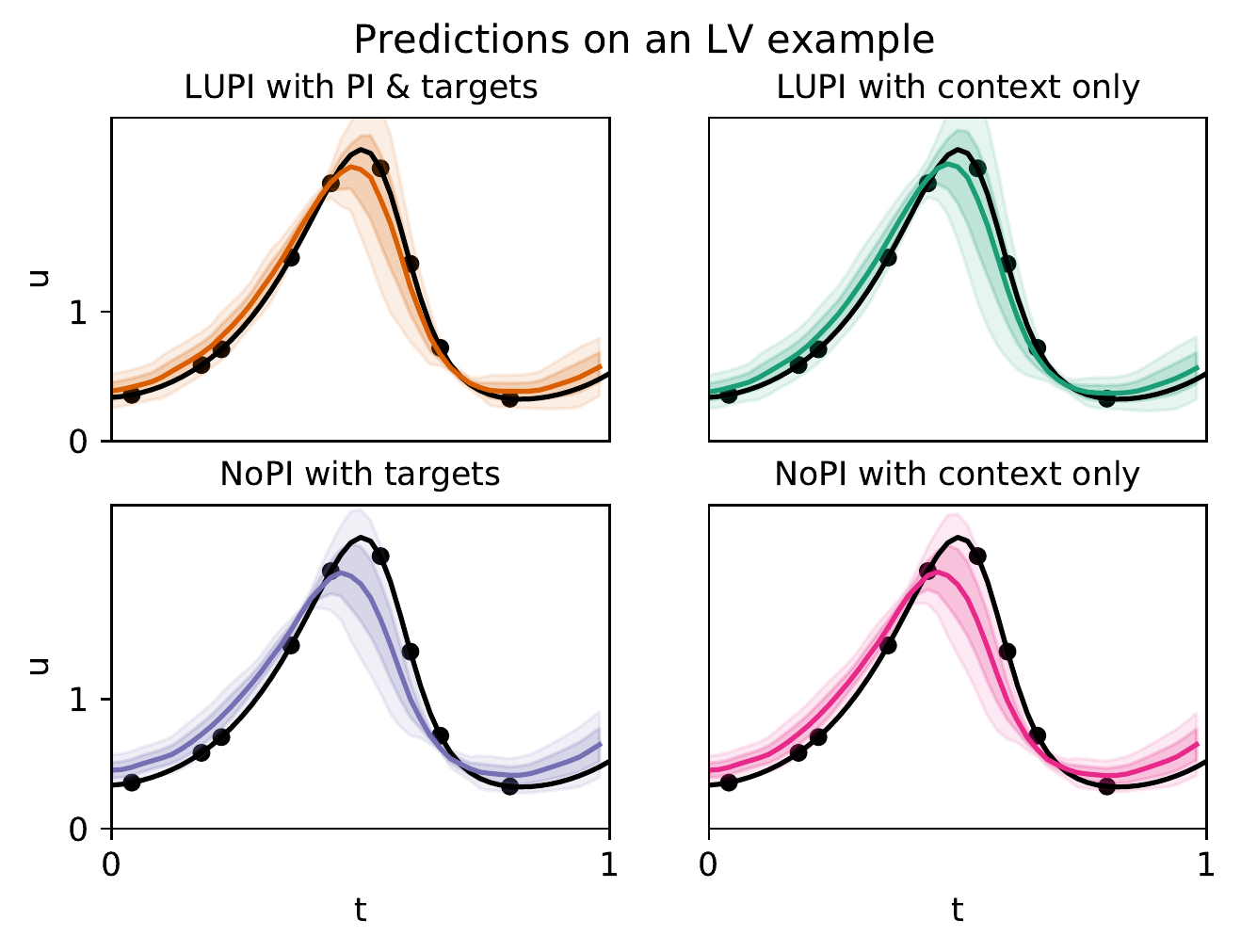}
         \caption{Example predictions}
         \label{fig:lv_predictions}
     \end{subfigure}
     \hfill
     \begin{subfigure}[b]{0.32\textwidth}
         \centering
         \includegraphics[trim =  5 5 5 21, clip=true, width=\textwidth]{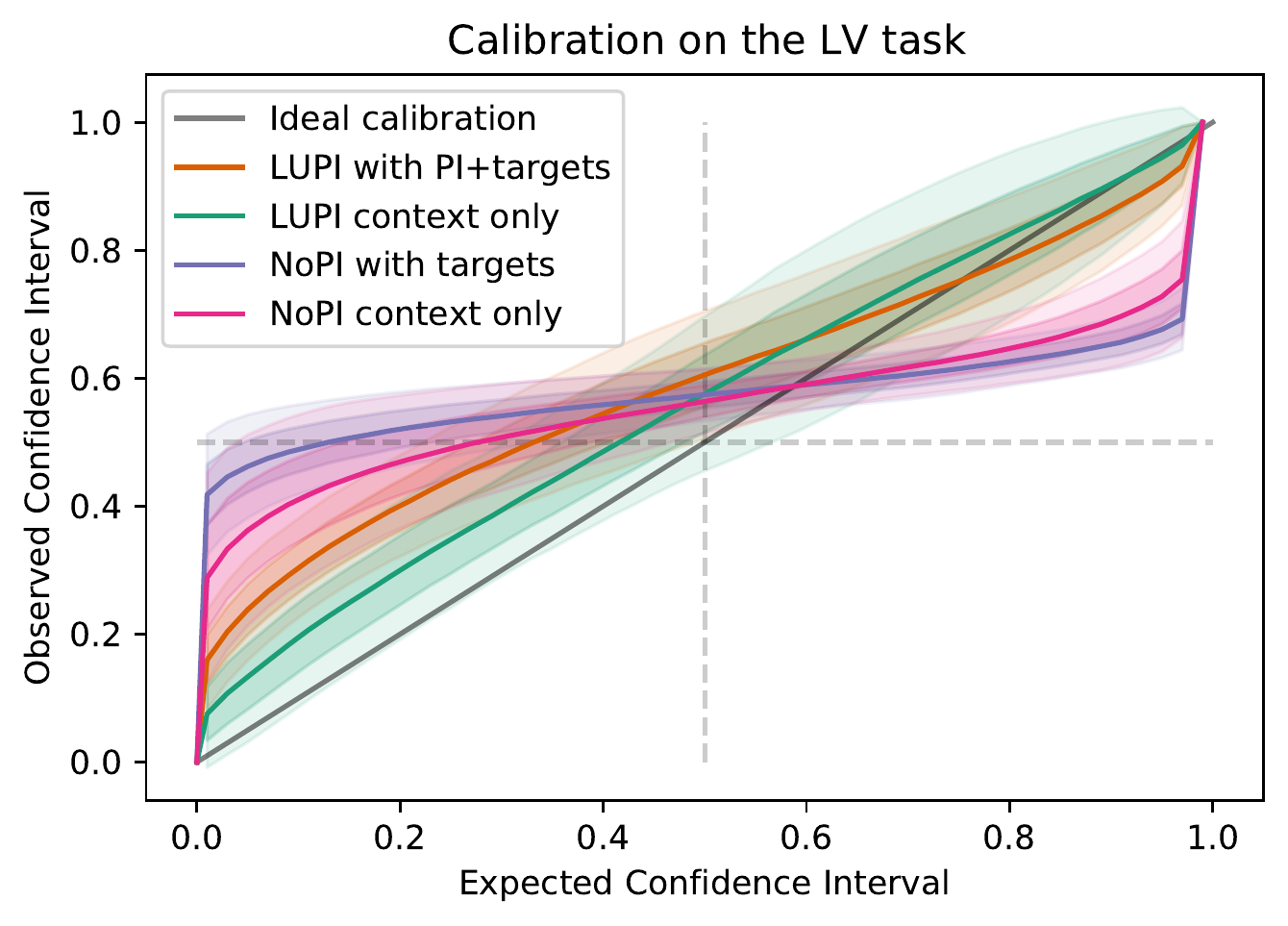}
         \caption{Calibration}
         \label{fig:lv_calibration}
     \end{subfigure}
     \begin{subfigure}[b]{0.32\textwidth}
         \centering
         \includegraphics[trim = 5 5 5 21, clip=true, width=\textwidth]{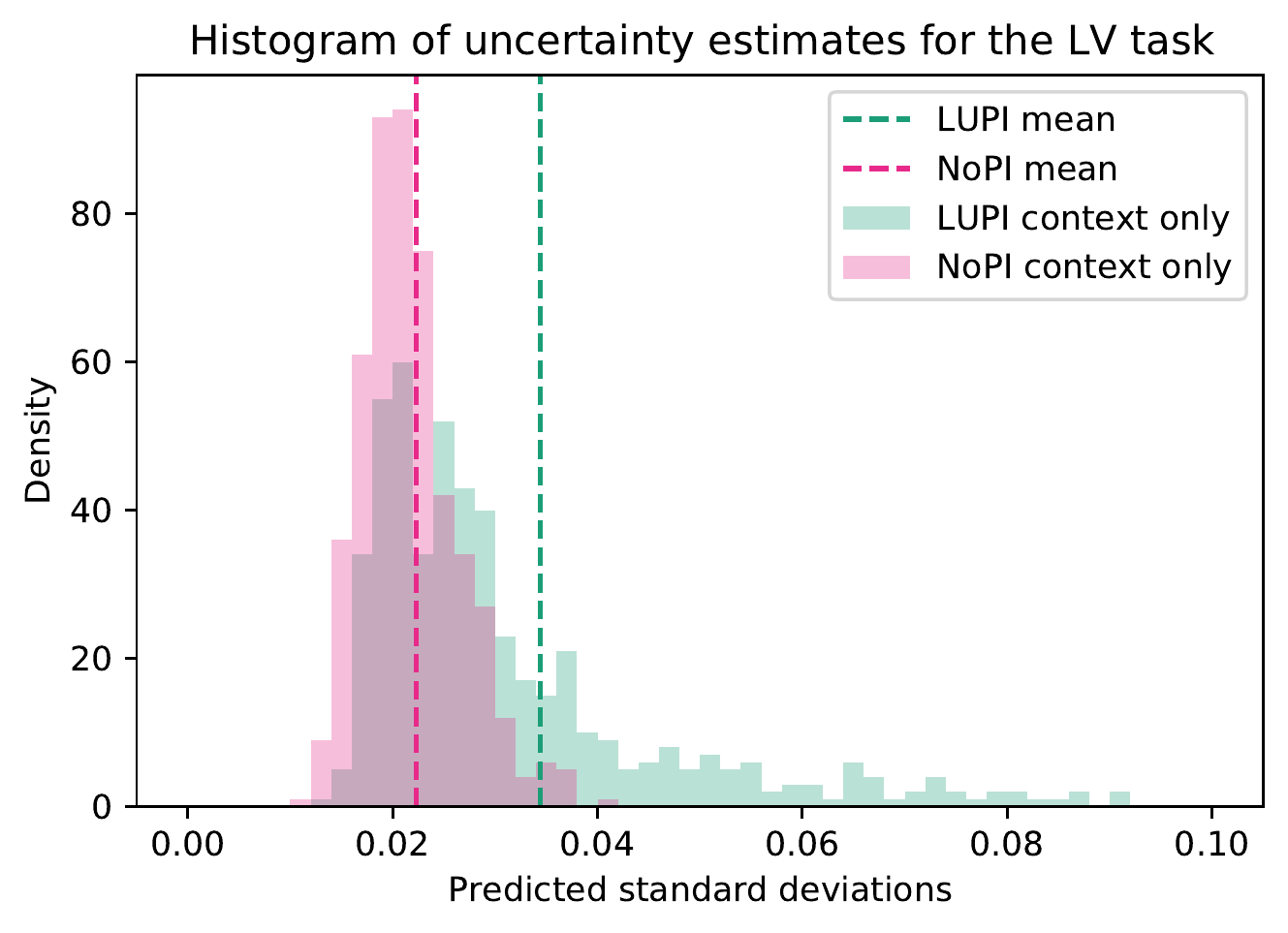}
         \caption{Sharpness}
         \label{fig:lv_sharpness}
     \end{subfigure}
        \caption{Various comparisons of trained LUPI and NoPI models on the Lotka-Volterra task with initial conditions, $(u(0),v(0))$. The LUPI model is more accurate and better calibrated, but less sharp. All the models are overconfident to some degree, but the LuPI models are significantly better calibrated.}
        \label{fig:three_plot_lv}
\end{figure}

Figure \ref{fig:three_plot_lv} shows additional details for the Lotka-Volterra experiment.

\end{document}